\documentclass[12pt]{article}
\usepackage[utf8]{inputenc}
\usepackage{graphicx}
\usepackage[margin=1in]{geometry}
\usepackage{url}
\usepackage{amsmath}
\usepackage{microtype}   
\usepackage{setspace}
\usepackage{todonotes}
\usepackage{booktabs}
\usepackage{siunitx}
\usepackage{xcolor}    
\usepackage{threeparttable}
\usepackage[hidelinks]{hyperref}
\graphicspath{ {figures/} }

\title{Climate Finance Bench}
\author{
  Rafik Mankour\thanks{Institut Louis Bachelier} \and
  Yassine Chafai$^*$ \and
  Hamada Saleh$^*$ \and
  Ghassen Ben Hassine$^*$ \and
  Thibaud Barreau$^*$ \and
  Peter Tankov\thanks{CREST, ENSAE, Institut Polytechnique de Paris and Institut Louis Bachelier}
}
\date{}

\begin{document}

\maketitle

\sloppy 
\begin{abstract}
\textbf{Climate Finance Bench} introduces an open benchmark that targets question–answering over corporate climate disclosures using Large Language Models. We curate 33 recent sustainability reports in English drawn from companies across all 11 GICS sectors and annotate 330 expert‑validated question–answer pairs that span pure extraction, numerical reasoning, and logical reasoning. Building on this dataset, we propose a comparison of RAG (retrieval-augmented generation) approaches. We show that the retriever’s ability to locate passages that actually contain the answer is the chief performance bottleneck. We further argue for transparent carbon reporting in AI‑for‑climate applications, highlighting advantages of techniques such as Weight Quantization.


\end{abstract}


\section{Introduction}

Climate finance aims to measure and control climate-related financial risks, to redirect financial flows towards the green sectors of the economy, and to provide incentives for the brown sectors to reduce their carbon footprint. All this requires reliable measurement of corporate climate risk exposures and environmental impacts.
Yet regulators and investors frequently cite a \emph{climate-data gap}: key indicators such as Scope 1–3 emissions, decarbonization targets or capital-expenditure plans are either missing, inconsistently reported or locked in unstructured PDF filings \cite{ngfs2022datagap}.

Climate-related financial disclosure has moved from a voluntary narrative exercise to a quasi-regulatory requirement.  Nineteen jurisdictions, including the EU, the United Kingdom and Japan, have either adopted or are piloting mandatory reporting regimes aligned with the \textit{Task Force on Climate-related Financial Disclosures} (TCFD) or the upcoming \textit{IFRS S2} and \textit{European CSRD} standards, covering more than 60\% of global GDP and most major capital markets\,\cite{fsb2023tcfd}. 

Credible transition plans, scenario-aligned metrics and financed-emission inventories are now essential inputs to prudential stress tests and capital allocation decisions \cite{bis2024climate,bundesbank2024data}. As a result, large corporates publish hundreds of pages of sustainability information, often in PDF
format, replete with tables, embedded figures and forward-looking statements. Central banks warn that robust, machine-readable climate data are now prerequisite for prudential stress testing and macro-financial stability.

Analysts aiming to extract information from corporate sustainability reports face three persistent bottlenecks:

\begin{enumerate}
  \item \textbf{Heterogeneity of formats.}  As there are no common reporting standards, disclosures vary widely in structure, terminology and granularity, hindering automated parsing and cross-company comparison.
  \item \textbf{Information overload.} Climate or ESG (Environmental, Social and Governance) reports can now exceed hundreds of pages and relevant data are scattered among other information, manual review at portfolio scale is no longer feasible.
  \item \textbf{Data quality and auditability.}  Stakeholders require exact numeric values (e.g.\ Scope 1 + 2 GHG emissions) and transparent provenance to guard against green-washing.
\end{enumerate}

\emph{Large Language Models} (LLMs) such as GPT-3~\cite{brown2020gpt3} and GPT-4~\cite{openai2023gpt4} can read long documents, extract key metrics and draft natural-language explanations.  However, vanilla LLMs suffer from two well-known limitations when applied to climate-finance documents: limited context windows that truncate relevant evidence, and a tendency to hallucinate plausible-sounding but unsupported figures~\cite{ji2023hallucinations}.
\emph{Retrieval-Augmented Generation} (RAG) addresses both issues by (1)~retrieving the most relevant text passages from the source documents and (2)~conditioning the LLM on those passages so that answers remain grounded in the original evidence~\cite{lewis2020rag,izacard2021fid}.

In this work we systematically evaluate RAG pipelines for climate-finance question answering, making the following contributions:

\begin{itemize}
    \item We establish a benchmark dataset comprising 33 climate disclosures, 330 analyst-curated questions and expert-validated reference answers that span data extraction, numerical reasoning and logical inference.
    \item We compare multiple RAG configurations: minimal dense retrieval, hybrid dense + BM25 and reranking, across several LLM back-ends, quantization levels and prompt strategies.
    \item We provide an open, reproducible test-bed that reports both answer accuracy and the \emph{carbon footprint} of each configuration, so that future research can optimize statistical as well as environmental   efficiency.
\end{itemize}

We assemble a representative sample of reports across all 11 GICS sectors.
By emphasizing factual correctness, including exact numbers and units, and by measuring retrieval coverage, we aim to help both researchers and practitioners develop trustworthy, resource-efficient question-answering systems for the climate-finance domain.

\section{Literature Review}


The rapid commercial rollout of large language models has brought an equally rapid migration toward \emph{retrieval-augmented generation}.  In less than two years, RAG has moved from a research curiosity to the backbone of production systems such as Bing Chat, Perplexity.ai and the many domain-specific applications built with LangChain or LlamaIndex.  Because it \emph{grounds} outputs in verifiable documents and avoids expensive model fine-tuning, RAG is now the de-facto recipe for deploying LLMs in high-stakes settings ranging
from medical question answering to legal contract analysis \cite{mialon2023survey,guo2023survey}.  The climate-finance domain is no exception: analysts need fact-checked answers, regulators demand provenance, and datasets change too frequently for parametric retraining alone.  Against this backdrop, we summarize below the core architectural choices in modern RAG pipelines and how they inform our benchmark design.

\paragraph{Retrieval-Augmented Generation (RAG).}
Early RAG systems \cite{lewis2020rag} combine a {dense} bi-encoder retriever with a seq2seq generator, injecting the top-$k$ retrieved passages into the generation context so that answers remain {grounded} in source documents.
Subsequent work shows that feeding dozens of passages and letting the decoder {fuse} evidence (the FiD architecture) yields large gains on knowledge-intensive tasks \cite{izacard2021fid}.
Best-practice surveys now recommend a \emph{hybrid pipeline} (dense embeddings $+$ BM25) followed by a cross-encoder reranker for precision, plus prompt instructions that discourage hallucination \cite{ragpractices}.
In short:
\begin{itemize}
   \item dense retrieval captures semantic paraphrases,
   \item sparse lexical search excels at exact figures and units, and
   \item reranking re-scores candidates jointly with the query to identify the truly relevant ones.
\end{itemize}
The generator is then tasked to quote or cite its supporting text, a practice that
empirically reduces unsupported claims and eases human verification.

Open-source efforts such as FinGPT \cite{yang2023fingpt} and ESG-BERT \cite{mehta2022esgbert} also show promise in adapting general-purpose LLMs to domain-specific tasks like ESG disclosure classification and financial forecasting. We believe that climate-finance QA tasks would benefit from similar fine-tuning initiatives.

The FinanceBench paper \cite{financebench} introduced a finance-specific benchmark to test LLMs in a question-answering setting. Their dataset covered a selection of 10\,000 questions across 40 publicly traded US companies from various sectors, drawn from real financial documents such as 10-K, 10-Q and 8-K filings, and they did a human evaluation over a subset of 150 evaluation questions. Although FinanceBench aimed to provide a broad finance-oriented benchmark, its reported performance evaluations were done primarily via human annotations, with a relatively permissive notion of “correctness” (e.g., minor deviations in units were not treated as fully incorrect) to ensure a good-faith understanding of the capabilities of the models.

Moreover, FinanceBench explored several LLM configurations, including “open book,” “closed book,” “retrieval,” and “long context.” Among these, only the “retrieval” setting fully aligns with RAG-based question answering. Inspired by the FinanceBench methodology, we focus here on climate-related disclosures, adopting and extending the retrieval pipeline to better handle complex disclosures such as tables and figures. We also incorporate insights from best-practice studies on retrieval-augmented generation \cite{ragpractices} to refine our approach.

In addition to FinanceBench \cite{financebench}, several other benchmarks focus on numerical and reasoning-based question answering from financial documents. FinQA \cite{chen2021finqa}, ConvFinQA \cite{chen2022convfinqa} and TAT-QA \cite{zhu2021tatqa} emphasize numerical reasoning and hybrid table-text data. These datasets are relevant for evaluating climate-finance models, especially those requiring computation over reported KPIs.

Within the ESG and climate disclosure space, Climate-FEVER \cite{diggelmann2020climatefever} and ClimRetrieve \cite{climretrieve} provide valuable baselines for factual verification and information retrieval respectively. While Climate-FEVER focuses on verifying real-world claims, ClimRetrieve offers climate-specific documents and retrieval annotations.

More recently, the Golden Touchstone benchmark \cite{wu2024touchstone} provided a bilingual and comprehensive evaluation framework for finance-specific LLMs and BloombergGPT \cite{shen2023bloomberggpt} demonstrated how domain pretraining boosts performance in financial QA.

\subsubsection*{Limitations of current methodologies and proposed improvements}
\vspace{-0.4em}
\paragraph{Dataset scope.}
Current methodologies evaluate sophisticated {reasoning} but often provide pre-extracted snippets for each question, leaving the retrieval challenge unsolved.
Full-document retrieval often covers only U.S.\ SEC filings or a limited set of sustainability reports.

\paragraph{Retrieval transparency.}
Few benchmarks disclose the exact retrieval pipeline used in baselines; hybrid search, cross-encoder reranking and document preprocessing are rarely benchmarked side-by-side, hindering reproducibility and progress measurement.

\paragraph{Evaluation methodology.}
Some benchmarks label an answer as correct only when the exact text string matches the reference answer. The challenge is to avoid letting hallucinations slip through without penalizing semantically correct answers written with different wording, units or abbreviations.
A fairer scheme should (i) award partial credit to answers that cover only part of the required information while taking into account their incompleteness and (ii) scale to large datasets.  
We therefore adopt automated grading with an LLM-as-a-Judge, while keeping a human-in-the-loop design.



\paragraph{Our contribution.}
\textbf{Climate Finance Bench} fills these gaps by:
\begin{enumerate}
  \item using {complete} ESG and climate reports (33 documents across all 11 GICS sectors), thereby testing end-to-end retrieval on long, heterogeneous PDFs;
  \item releasing a reference \emph{hybrid + reranking} pipeline so that future work has a documented, strong baseline to iterate on;
  \item enforcing a non-binary, 3-point grading scheme based on LLM-as-a-Judge;
  
  
  \item establishing estimates for the carbon emissions associated to the tools we experiment on.
\end{enumerate}
By unifying retrieval, reasoning and environmental accountability, our benchmark aims for trustworthy, resource-efficient question answering in the climate-finance domain.

\section{Methods}

\subsection{Data Collection and Curation}

\subsubsection{Selection of Reports.}
We gathered 33 climate reports from large publicly traded companies spanning multiple regions (e.g., CAC40 and DAX40 in continental Europe, FTSE in the UK, S\&P500 in the US) and covering all 11 GICS sectors. We ensured:
\begin{itemize}
    \item At least one company from Communication Services, Real Estate and Health Care sectors.
    \item At least two companies from each of the 8 remaining sectors, preferably from different sub-sectors.
    \item Exactly one recent climate or sustainability report per company, capturing the latest relevant fiscal year.
\end{itemize}
The full list of selected companies is available in Appendix C.

\subsubsection{Question Formulation and Annotation.}
We relied on ESG experts to provide 10 questions per report. The questions reflect two modalities (metric-related and domain-related) and three categories:
\begin{enumerate}
    \item \textbf{Pure Extraction}: directly retrieve facts.\\
          \textit{Example: Has the company identified significant decarbonization levers? If yes, detail them.}

    \item \textbf{Numerical Reasoning}: extract figures and/or perform calculations.\\
          \textit{Example: What is the company's carbon intensity (in tCO\textsubscript{2}\,/\,million USD) for FY 2023?  
          If not reported, compute it by dividing total carbon emissions by that year’s revenue.}

    \item \textbf{Logical Reasoning}: combine multiple data points to infer an answer.\\
          \textit{Example: Does the company have a decarbonization trajectory compatible with a 1.5 °C or 2 °C scenario?}
\end{enumerate}

Seven analysts carried out the following steps:
\begin{itemize}
    \item Carefully read the assigned climate report(s).
    \item Provide written, reference answers (the ``Gold Standard'') for each of the 10 questions, along with document excerpts, page numbers and an indication of whether the relevant information was found in text, a table, or a figure.
    \item Follow a unified annotation guide to ensure consistent handling of numerical values, units and references. An adapted version of this guide is available in Appendix D for reference.
\end{itemize}
Two ESG domain experts resolved ambiguous cases and a final quality-control check was performed to confirm the validity of all annotations. Ultimately, we obtained 330 question--answer pairs.

\subsubsection{Resulting Dataset Structure.}
We store our dataset in a table of 330 rows and 13 columns. Each row corresponds to a single question about a specific company’s report, containing:
\begin{itemize}
    \item \texttt{Company name} and \texttt{Fiscal year}.
    \item \texttt{Question ID}, \texttt{Question text}, \texttt{Type of question}.
    \item \texttt{Answer} (reference/Gold Standard).
    \item \texttt{Documents}, \texttt{Pages}, \texttt{Document extracts}, \texttt{Extract type}.
\end{itemize}

\subsection{Hardware and Environment}

We conducted the experiments primarily on a GPU environment available through Kaggle (Notebooks hosted on GCP). This environment provides:
\begin{itemize}
    \item 60 GB of storage and 30 GB of RAM.
    \item Access to a GPU P100 (16 GB memory).
\end{itemize}

For LLMs with publicly available APIs (Claude Sonnet and GPT-4o), as well as Qwen2.5 and DeepSeek R1 via Nebius’ API, calls were invoked directly from the Kaggle notebook.

\subsection{Data Extraction and Chunking}

Our RAG pipeline builds on the foundational architecture introduced by Lewis et al. \cite{lewis2020rag}, while incorporating best practices from the FiD architecture \cite{izacard2021fid} and Atlas model \cite{izacard2022atlas}. These methods highlight the importance of fusing retrieved passages in the decoder and optimizing passage selection for generation.

In designing the retrieval system, we consulted evaluations from the KILT \cite{petroni2021kilt} and BEIR \cite{thakur2021beir} benchmarks, which stress the need for robust lexical-dense hybrid retrieval and reproducible metrics across knowledge-intensive tasks.

\paragraph{PDF Extraction.}
We experimented with two main approaches:
\begin{itemize}
    \item \textbf{LangChain loaders} that use the Unstructured library, providing a quick way to parse text.


    \item \textbf{Docling}, which converts PDF files to HTML/Markdown to preserve more structure (e.g., tables, figure captions).
\end{itemize}
Although Docling can better retain table formatting, it sometimes introduces noise such as HTML tags, which can complicate retrieval.

\paragraph{Chunking Strategy.}
To avoid having either overly large text chunks or fragments broken mid-table, we used:
\begin{itemize}
    \item A base chunk size of 2048 tokens, with an overlap of 204 tokens (i.e., 10\%).
    \item Logic to avoid breaking tables and to merge small paragraphs with relevant headings.
\end{itemize}

\subsection{Vector Indexing and Retrieval}

We vectorized all chunks using \texttt{sentence-transformers/all-mpnet-base-v2} and stored them in a FAISS index for nearest-neighbor retrieval. We explored two RAG retrieval configurations:

\begin{itemize}
    \item \textbf{Minimal}: return the top $k$ = 12 chunks based solely on cosine similarity scores with the question embedding.
    \item \textbf{Best Practices}: a “hybrid” approach combining:
    \begin{enumerate}
        \item \textbf{Semantic and lexical retrieval}: weighted combination of top results from a dense (semantic) retriever (75\%) and BM25 (25\%).
        \item \textbf{Fusion and Reranking}: we first fuse the top 20 chunks using Reciprocal Rank Fusion, select 8 best chunks, then apply a cross-encoder reranker to the next 12 to pick 4 additional relevant chunks, for a total of 12.
    \end{enumerate}
\end{itemize}

\subsection{LLMs for Generation}

We tested five LLMs under both minimal and hybrid retrieval:
\begin{itemize}
    \item \textbf{Llama3.1 8B Instruct}\footnote{\url{https://huggingface.co/meta-llama/Llama-3.1-8B-Instruct}} and \textbf{Llama3.1 8B Instruct quantized in 4-bit}.
    \item \textbf{Mistral Nemo Instruct 12B\footnote{\url{https://huggingface.co/mistralai/Mistral-Nemo-Instruct-2407}} quantized in 4-bit}.
    \item \textbf{Claude Sonnet 3.5 2024-06-20}\footnote{\url{https://www.anthropic.com/index/claude-3-family}}.
    \item \textbf{GPT-4o}\footnote{\url{https://openai.com/index/hello-gpt-4o}}.
\end{itemize}

Two more LLMs were tested under hybrid retrieval only:
\begin{itemize}
\item \textbf{Qwen2.5-72B}\footnote{\url{https://huggingface.co/Qwen/Qwen2.5-72B}}.
\item \textbf{DeepSeek R1}\footnote{\url{https://arxiv.org/abs/2501.12948}}.
\end{itemize}

We set the temperature to 0.2 (for more deterministic answers) and limited the maximum output to 512 tokens to avoid overly long responses. All queries shared the same two–part prompt shown below, with a system prompt instructing the LLM to answer strictly based on the retrieved chunks, in order to refrain from hallucinating data. 

Curly-brace placeholders (\texttt{\{company\}}, \texttt{\{context\}}, \texttt{\{question\}}) are filled at run-time:

\begin{quote}\small
\begin{verbatim}
----- System prompt (fixed) -----
You are a documentary assistant.
Answer the question about the mentioned company based on the
provided context that was extracted from climate or sustainability
reports. Do not add any additional notes.
If the answer to the question is missing from the provided context
and you cannot conclude on it on your own, indicate this sincerely.

Here are three examples of the format to follow in your reply:
###
Human: Does the company have a climate change mitigation objective 
for FY2023?
AI: Yes, the company aims to become net zero by 2030 on its Scope 
1, 2 and 3 emissions.

Human: Does the company have a climate change mitigation objective 
for FY2023?
AI: No, the company clarifies its intention not to pursue a 
net-zero target.

Human: Does the company disclose a Transition Plan for FY2023? 
If yes, highlight its main characteristics.
AI: Not available in the retrieved information.
###

----- User prompt (filled per query) -----
Here are excerpts from documents about the company {company}:
###
{context}
###
Here's the question asked by the user:
Question: <<< {question} >>>
\end{verbatim}
\end{quote}

To compare full‑precision with compressed models under identical conditions and to fit \textit{Mistral Nemo Instruct 12B} on the 16GB GPU available in our local environment, we performed post‑training, 4‑bit weight quantization with the \texttt{bitsandbytes} library (v0.43). Both \textit{Llama3.1 8B Instruct} and \textit{Nemo Instruct 12B} were loaded and no additional fine‑tuning was applied. We evaluated the quantized checkpoints in inference‑only mode, using exactly the same decoding hyper‑parameters (temperature 0.2, top‑p 0.95, \texttt{max\_new\_tokens}=512) as for the baselines.  Quantization reduced resident GPU memory from 16GB to 6GB for \textit{Llama3.1 8B Instruct} and from 24GB to 9GB for \textit{Nemo Instruct 12B}, enabling local execution.

\section{Results}

\subsection{Manual Evaluation and LLM-as-a-Judge}

We first performed a human evaluation of RAG outputs. Human annotators labeled each answer as either \textit{correct}, \textit{incomplete}, or \textit{incorrect}, with exactness in numeric values and appropriate textual evidence being key factors. However, human evaluation is time-consuming and can be subjective.

To reduce manual overhead, we tested an LLM-based grader (\textit{LLM-as-a-Judge}) that compares each RAG-generated answer to the reference answer, factoring in the context of the original question. We found the highest alignment with human judgments (over 80\% agreement) when using Claude with the question prompt included. Thus, for the main experiments, we rely on the LLM-as-a-Judge framework for scalable evaluation. 

A full description of the LLM-as-a-Judge evaluation protocol is available in Appendix~\ref{app:llmjudge}.

\subsection{Comparing RAG Configurations and LLMs}
 
\subsubsection{Minimal RAG.}
\begin{figure}[htbp]
    \centering      
    \includegraphics[width=\linewidth]{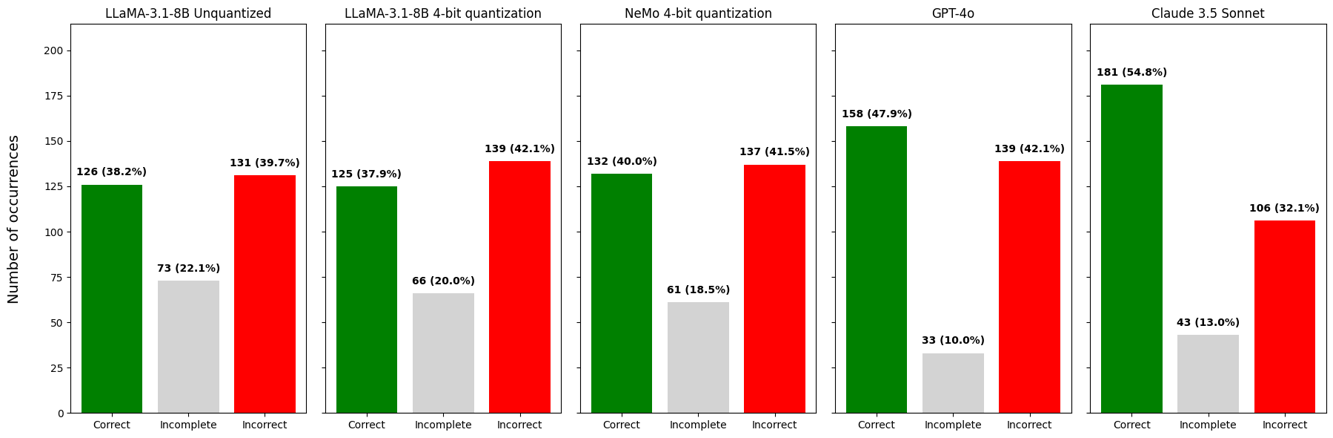}

    \caption{%
        Accuracy breakdown (\textit{correct}, \textit{incomplete}, \textit{incorrect})
        for the \textbf{Minimal RAG} configuration across five LLMs.
        \newline
        }
    \label{fig:minimal-rag-performance}
\end{figure}

Under the minimal retrieval approach, smaller or quantized LLMs (Llama3 8B Instruct, Llama3 Instruct 4-bit, Nemo Instruct 4-bit) achieved 35--40\% correct responses. Larger models (Claude and GPT-4o) performed significantly better, around 50--55\% correct. As shown in Figure~\ref{fig:minimal-rag-performance}, GPT-4o and Claude 3.5 outperform
smaller 4-bit models by roughly 15 percentage points under the minimal setting.

\subsubsection{Hybrid RAG (BM25 + Reranking).}
\begin{figure}[htbp]
    \centering
    \includegraphics[width=\linewidth]{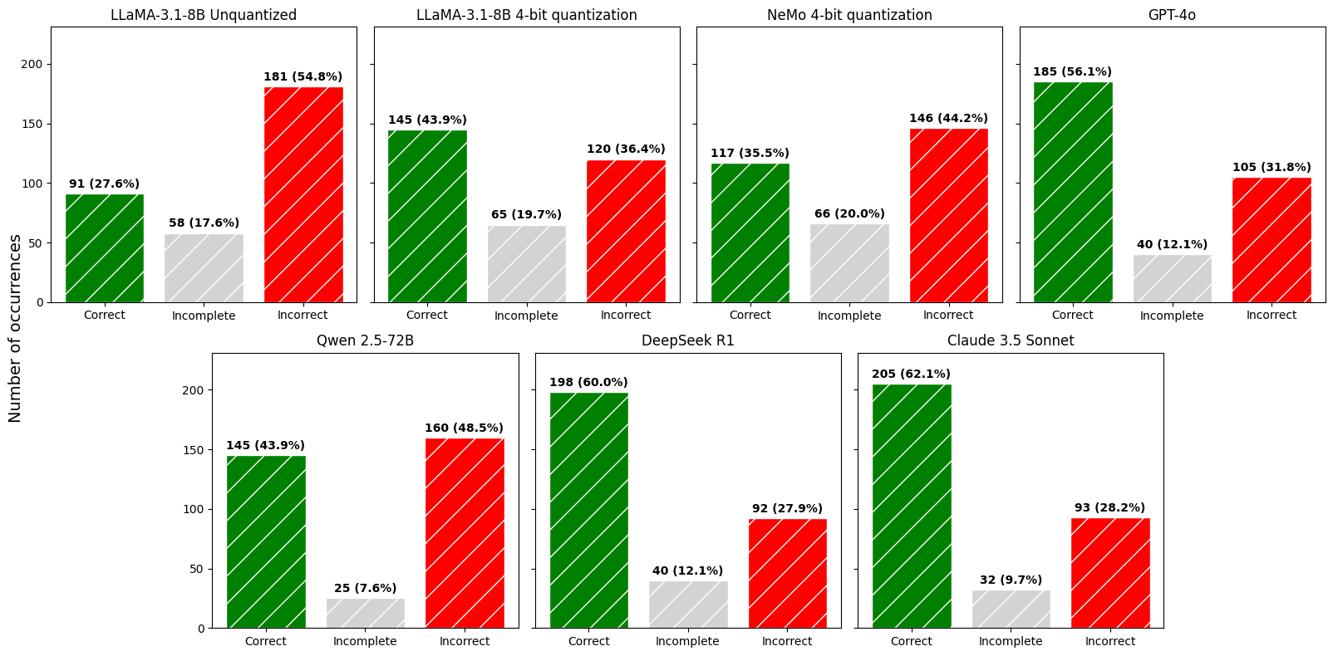}

    \caption{Accuracy breakdown (\textit{correct}, \textit{incomplete}, \textit{incorrect})
             for the \textbf{Hybrid RAG} configuration across the seven LLMs tested.}
    \label{fig:hybrid-rag-performance}
\end{figure}

Adding BM25 lexical retrieval and reranking yielded substantial gains for larger LLMs. As illustrated in Figure~\ref{fig:hybrid-rag-performance}, Claude 3.5 remains in the lead at 62\%, but the performance gap has narrowed: DeepSeek R1 attains 60\%, while Qwen2.5-72B reaches 44\%.  In other words, switching from Claude 3.5 to DeepSeek R1 incurs an $\approx 2$pp drop, even though Claude consumed about five times less output tokens per answer on average.
This suggests that retrieval quality, rather than model capacity, seems to be the current bottleneck in our experiments. 

\subsubsection{Minimal vs.\ Hybrid.}
\begin{figure}[htbp]
    \centering
    \includegraphics[width=\linewidth]{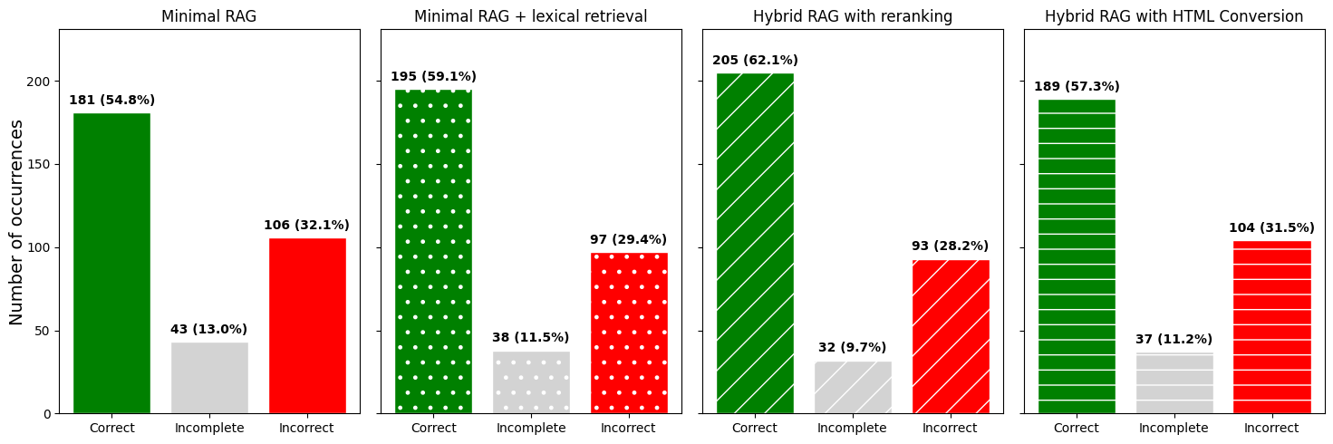}
    \caption{%
      Stepwise impact of successive retrieval upgrades on answer quality
      (Minimal RAG  →  + BM25 lexical  →  + reranking  →  + HTML conversion).
      Adding BM25 improves the correct–answer rate from 54.8\% to 59.1\%,
      and the hybrid dense–sparse \& reranking scheme lifts it further to
      62.1\%.  Introducing Docling’s HTML conversion {without extra
      post‑processing} brings the score down to 57.3\%, indicating that
      raw structural noise can offset earlier gains.  Bars show absolute
      counts (annotated) and the associated share of the 330‑question test
      set.%
    }
    \label{fig:retrieval-step-effect}
\end{figure}

Hybrid retrieval helps large instruction-tuned models but can hurt smaller or highly-quantized ones.  Qwen2.5-72B and DeepSeek R1 were not run under minimal retrieval, so direct deltas are unavailable.  Incremental improvements on the generation side alone are unlikely to break the 65\% barrier without a better retriever. Figure \ref{fig:retrieval-step-effect} details these jumps: BM25 adds +4.3pp, reranking another +3.0pp, while an un‑filtered HTML conversion costs 4.8pp.

\subsubsection{Effect of Docling Conversion.}
While Docling potentially preserves the layout of tables and figures, we observed a performance decrease of several percentage points, likely due to extra HTML tokens and parsing noise degrading retrieval's performance. This suggests that more advanced post-processing might be required to capitalize on Docling’s structural advantages.

\subsubsection{Quantized vs. Full-Precision.}
\begin{figure}[htbp]
    \centering
    \includegraphics[width=\linewidth]{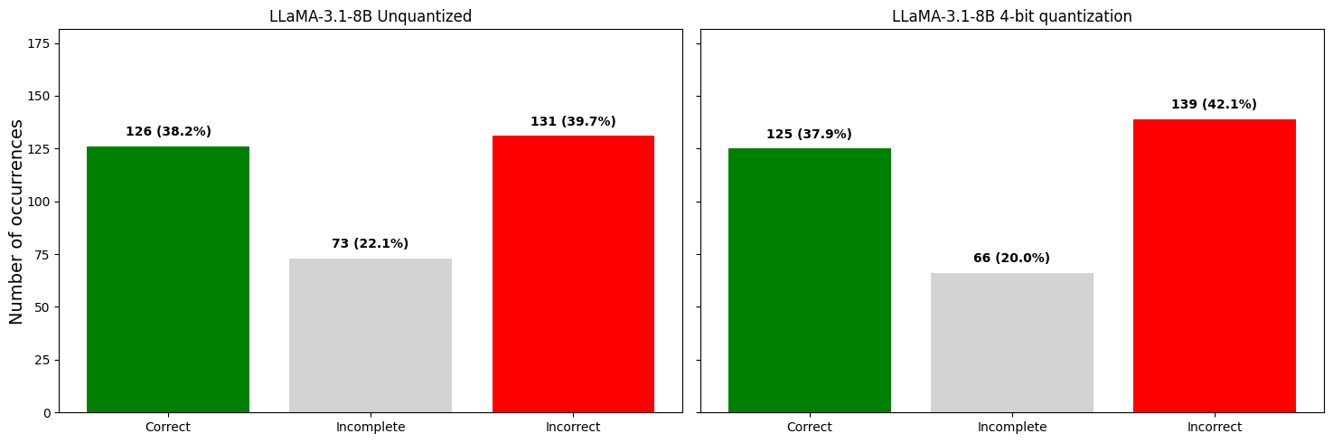}
    \caption{%
    Comparison of \textbf{LLaMA 3.1-8B Unquantized} and \textbf{4-bit Quantized} under the minimal RAG setting.
    Quantization leads to negligible accuracy loss while significantly reducing resource usage.
}
    \label{fig:quantized_vs_unquantized_llama3}
\end{figure}

For smaller-scale Llama models, we found that 4-bit quantization introduced only minor accuracy differences (within 1--2\%), while significantly reducing memory usage and energy consumption.
Figure~\ref{fig:quantized_vs_unquantized_llama3} illustrates this comparison for LLaMA-3.1 8B, where the unquantized and quantized versions perform similarly across all three correctness classes.
In resource-constrained settings, 4-bit quantization is therefore a compelling strategy to lower computational cost and associated carbon emissions without critically hurting correctness.

\subsection{Performance by Question Type}

\begin{figure}[htbp]
    \centering
    \includegraphics[width=\linewidth]{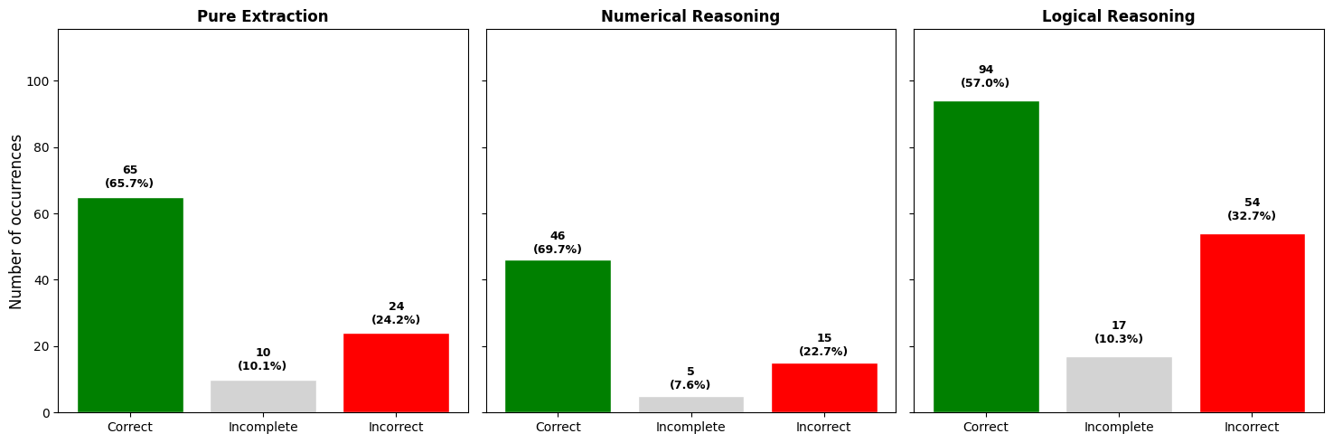}
    \caption{%
    Break‑down of answer quality for each question category under the best‑performing setup (Claude 3.5 + hybrid retrieval).  Numerical reasoning edges out pure extraction, while logical reasoning lags behind because it demands multi‑hop synthesis across passages.
    }
    \label{fig:question-type-performance}
\end{figure}

Our question set includes extraction, numerical reasoning and logical reasoning categories, each varying in difficulty. Figure~\ref{fig:question-type-performance} reveals patterns that are not obvious from aggregate accuracy alone. 

In our best configuration (Claude 3.5 + Hybrid retrieval), numerical reasoning questions had the highest correct-answer rate, when one might expect numerical reasoning to be harder than pure extraction (69.7\% vs.\ 65.7\% correct). A qualitative error analysis indicates why: in some of the numerical questions, the relevant arithmetic had already been carried out in the source document, so the task collapses into precise retrieval plus unit normalization.

Logical questions expose a retrieval bottleneck as logical‑reasoning items require chaining multiple facts scattered across a report. Broad or ambiguous queries (e.g., “Which topics have been assessed to be material?”) caused more errors or incomplete answers. 

Properly structured queries, with clear numeric or factual targets, were most reliably answered. Future iterations of \textit{Climate Finance Bench} will therefore log intermediate reasoning traces to diagnose whether errors originate from retrieval omissions or reasoning failures and to build custom processes per question in order to improve the success rate of answers.

\subsection{GHG Emissions and Environmental Footprint}

Measuring emissions from AI usage aligns with growing interest in sustainable ML practices. Patterson et al. \cite{patterson2022carbon} provide methodology to estimate the carbon impact of training large neural networks, while Schwartz et al. \cite{schwartz2019green} advocate for “Green AI,” urging the community to prioritize energy efficiency in both training and deployment.

In keeping with the climate focus, we also conducted a rough measurement of the greenhouse gas (GHG) emissions attributed to these experiments:
\begin{itemize}
    \item For Claude and GPT-4o, we estimated an upper bound per query using external providers like \textit{EcoLogits}, then generalized across all runs.
    \item For local or Kaggle-based models, we approximated total GPU usage with \textit{CodeCarbon} logs, dividing by the total number of queries to compute a per-query footprint.
    \item We could not obtain estimates for Qwen and DeepSeek as they were called through Nebius's API.
\end{itemize}

Aggregating CPU, memory and GPU energy logs, \textit{CodeCarbon} estimates for local runs and per‐query upper–bound figures from \textit{EcoLogits} for the proprietary API calls, we estimated the GHG footprint of each model with a confidence interval due to the wide uncertainty bands published for OpenAI’s and Anthropic’s (Claude) back‑end infrastructure. 

Table~\ref{tab:emissions_per_query} provides emissions per question answered for each model.
\newpage

\begin{table}[h]
\centering
\begin{tabular}{lcc}
\toprule
\textbf{Model / configuration} & {\textbf{Emissions} (g\,CO\textsubscript{2}\,eq / query)} & {\textbf{Uncertainty} (g\,CO\textsubscript{2}\,eq / query)} \\ 
\midrule
LLaMA 3.1-8B (full precision)  & 2.79 & -- \\[2pt]
LLaMA 3.1-8B (4-bit)           & 0.70 & -- \\[2pt]
Mistral NeMo-12B (4-bit)       & 1.15 & -- \\[2pt]
GPT-4o (API)                   & 7.18 & ±\,4.00 \\[2pt]
Claude 3.5 Sonnet (API)        & 8.15 & ±\,4.52 \\[2pt]
Vector-store build\textsuperscript{*} & 0.30 & n/a \\
\bottomrule
\end{tabular}
\caption{Average GHG emissions per query.
The asterisk marks a one-off indexing cost if diluted over 330 queries, shown for scale.}
\label{tab:emissions_per_query}
\end{table}

Two points emerge:
(i)~\textbf{API calls dominate}: GPT-4o or Claude generate higher emissions than smaller models, disproportionately to the improvement in performance;
(ii)~\textbf{Quantitation pays off}: the 4‑bit quantization of Llama3.1 8B reduced emissions by a factor of~$\approx4$, yielding an emissions saving of roughly \(75\%\) with negligible accuracy loss. Retrieval and indexing remain almost negligible by comparison. 

This breakdown illustrates why future work should favor lightweight local models to factor in environmental impacts, whenever accuracy allows, and report vendor-side carbon metrics more transparently if practitioners are to measure carbon impacts with tighter bounds.

A full description of the emissions estimation protocol is available in Appendix~\ref{app:emissions}.

\section{Limitations}

This first release of \textit{Climate Finance Bench} covers only 33 sustainability reports, mostly large‑capitalisation firms headquartered in Europe or North America and is therefore not representative of emerging‑market issuers, small and medium enterprises, or non‑English filings.  

Despite a two-step review process, Gold Answers for multi‑step reasoning items still contain a degree of subjectivity, which may propagate noise into model‑versus‑human comparisons. 

Because climate disclosures are largely self‑reported, any inaccuracy or greenwashing in the underlying documents can bias both retrieval and evaluation.

Sector-wide queries based on information spanning multiple companies have not been considered yet.

Finally, the benchmark depends on PDF extraction and English‑language processing. Heterogeneous web formats and multilingual reports, which are common in practice, remain outside the current scope and should be addressed in future iterations.

\section{Conclusion and Practical Implications}
\label{sec:conclusion}

Our strongest configuration (Claude 3.5 combined with hybrid dense sparse retrieval and cross-encoder reranking) answered 62\% of the 330 question benchmark correctly and a further 10\% answers were partially incomplete.  Put differently, roughly three-quarters of the outputs were at least directionally useful, while one quarter remained factually wrong or unsupported.

\paragraph{What these numbers mean in practice.}
\begin{itemize}
  \item \textbf{Augmented analyst workflows.}  At 62 \%+10 \% accuracy, a RAG pipeline can already serve as first-pass summarization or evidence‐surfacing tools. In a typical ESG due-diligence loop, analysts spend the majority of their time locating passages, tables and footnotes. Automating that step can cut reading time even when every answer is manually verified afterwards.
  \item \textbf{Human-in-the-loop is still mandatory.} A 25-30 \% error rate remains too high for regulatory disclosure, portfolio weighting or automated sustainability scoring.  Every generated answer therefore needs stronger safeguards against hallucination and a review layer. In practice, showing the retrieved snippets alongside the model’s answer enables an experienced analyst to validate or override the response in a few seconds.
  \item \textbf{Retrieval dominates further gains.}  Numerical and logical questions fail largely because the correct passage never reaches the context window.  Hence marginal gains from scale (moving to larger proprietary models) are smaller than gains from smarter retrieval or domain-aware chunking.
  \item \textbf{Environmental trade-offs.}  Deploying Claude 3.5 or GPT-4o in production multiplies carbon emissions per query relative to a quantized local models. Organizations that prioritize sustainability can already choose lighter models with human review to keep both error rates and emissions within acceptable bounds.
\end{itemize}

The benchmark therefore positions current RAG technology not as a
replacement for ESG analysts but as a force multiplier that can save time, reduce tediousness and broaden coverage, while leaving final decision-making to human expertise.

\section{Moving Forward}

We plan to incorporate additional data sources to broaden coverage of companies, sectors and sustainability metrics.

We also anticipate exploring new retrieval strategies (e.g., domain-specific expansions, robust table extraction) and extended evaluation metrics beyond simple correctness, such as faithfulness to source and the ability to handle multi-document summaries.

A first key direction is the development of adaptive answering methods. By tailoring retrieval prompts and selection strategies to the type of question (e.g., numerical, logical, extractive), we can significantly reduce the semantic gap between query and document content. Typed-RAG, for instance, introduces a type-aware decomposition method that improves answer precision for complex question formats~\cite{lee2025typedrag}. Similarly, task-aware retrieval using instructions improves retrieval relevance and accuracy while reducing the need for model size scaling~\cite{asai2023taskaware}. These approaches allow RAG systems to better align context construction with task intent, ultimately improving factual accuracy and minimizing hallucination.

Another line of enhancement involves incorporating structured knowledge representations into the pipeline. We plan to prototype a GraphRAG variant, which extracts ESG-specific entities, numeric values and relations to construct a knowledge graph used during retrieval. This hybrid approach supports symbolic reasoning, enables multi-hop inference, and improves retrieval faithfulness~\cite{peng2024graphrag}. Recent results show that combining document graphs with entity-aware retrieval yields better coverage and more structured answers in focused summarization and QA tasks~\cite{edge2024graphragsumm}. This is especially relevant for climate disclosures, where facts are often interdependent and scattered across tables, figures and text.

To improve cost-efficiency and reduce emissions, we also aim to implement dynamic model routing. Inspired by cascade models and energy-aware inference pipelines~\cite{gupta2024cascades, maliakel2024energyllm}, this method routes queries to lightweight models for simple lookups and reserves larger models for complex or high-risk questions. In a benchmark context, this means using quantized models or symbolic solvers when appropriate, reducing unnecessary compute load while preserving accuracy. Dynamic routing aligns with our sustainability goals by lowering energy use and latency without sacrificing interpretability or correctness.

Finally, we plan to explore the Agentic RAG paradigm, which allows an orchestration agent to select and sequence retrieval tools at runtime. By combining document search, knowledge graph access, and symbolic modules under agentic control, the system can flexibly plan how to answer a question, rather than relying on a fixed RAG pipeline. The ReAct framework~\cite{yao2023react} and Toolformer~\cite{schick2023toolformer} demonstrate that language models equipped with action-selection capabilities can improve factual consistency, reduce hallucinations, and generate more interpretable reasoning traces. In our case, such orchestration can enhance trust in climate-finance QA outputs and ensure that high-stakes queries follow robust, auditable decision paths.

\section{Data accessibility}

We have made our dataset and code files available in a dedicated repository:

\begin{center}
\href{https://github.com/Pladifes/climate_finance_bench}
{\texttt{github.com/Pladifes/climate\_finance\_bench}}
\end{center}

All dataset assets (PDF excerpts, questions and gold answers) are distributed under the Creative Commons Attribution‑NonCommercial‑ShareAlike 4.0 International (CC‑BY‑NC‑SA 4.0) licence.

The repository includes notebooks so that users can:
\begin{itemize}
\item generate vector stores from reports and
\item run RAG pipelines by choosing from multiple retrieval and LLM options.
\end{itemize}

Users can experiment with various RAG configurations, test different LLMs and compare results.

\section{Acknowledgements}

We would like to thank the analysts of the Institut Louis Bachelier Labs for their availability in building and curating this benchmark:
\textbf{Pauline Aumard}, \textbf{Ana Vallejo}, \textbf{Adel Medjbari},  \textbf{Iker Tardio}, \textbf{Gabriel Levy}, and \textbf{Lina Jouker}.
Their question design and document‐level annotations were essential to the study.

We also thank \textbf{Yuri Vorontosov} (QuePasa) for his valuable recommendations and for providing compute resources that made these experiments possible.

\section{Funding}
This work was financed by Agence Nationale de Recherche via the Pladifes project (ANR-21-ESRE-0036). 

\newpage

\appendix
\section{Appendix A \\ Automated Grading with a LLM‑as‑a‑Judge}
\label{app:llmjudge}

\subsection{Human vs.\ Automatic Evaluation}

\paragraph{Human grading.}
For a sample of 330 RAG answers, each one was labeled by an annotator on a three‑level scale:
\begin{itemize}
  \item \textbf{Correct}: matches the gold answer within a narrow tolerance for wording;
  \item \textbf{Incomplete}: on the right track but missing key details;
  \item \textbf{Incorrect}: factually wrong or off‑topic.
\end{itemize}

\paragraph{LLM‑as‑a‑Judge.}
To scale evaluation we asked a LLM to grade the same answers.
Given the \emph{question}, the \emph{RAG answer} and the \emph{gold answer}, the LLM must output only a number corresponding to the labels:
\[
  2\;(\text{correct}),\quad
  1\;(\text{incomplete}),\quad
  0\;(\text{incorrect}),
\]
with no commentary.

\subsection{Prompt Design}

We tested three variants:
\begin{enumerate}
  \item \textbf{Llama3.1 8B Instruct}, \emph{without} a reminder of the question;
  \item \textbf{Llama3.1 8B Instruct}, \emph{with} a reminder of the question;
  \item \textbf{Claude 3.5 Sonnet}, with a reminder of the question.
\end{enumerate}

\subsection{Agreement with Human Judgements}

A \textit{soft} match counts \(\{\text{correct},\text{incomplete}\}\) as agreement, whereas a \textit{hard} match counts only \(\{\text{correct}\}\).

\begin{table}[ht]
  \centering
  \caption{Human–LLM agreement on the 330‑answer test set}
  \label{tab:llm_human_agreement}
  \begin{tabular}{lcc}
    \textbf{Grader setup} & \textbf{Soft match} & \textbf{Hard match} \\
    \hline
    Llama3.1 8B Instruct (no question) & 272/330 = 82.4\% & 165/330 = 50.0\% \\
    Llama3.1 8B Instruct (with question) & 280/330 = 84.4\% & 179/330 = 54.2\% \\
    Claude 3.5 Sonnet (with question) & 277/330 = 83.9\% & 227/330 = 68.7\% \\
    \hline
  \end{tabular}
\end{table}

Claude 3.5 achieved the highest hard‑match rate (68.7\%) while maintaining soft‑match parity with Llama3.1 8B Instruct, and is therefore used for all subsequent automatic evaluations.

\subsection{Distribution Shift}

Table \ref{fig:llm_judge_vs_human} contrasts the label distributions produced by human graders and by Claude 3.5.  
The LLM is noticeably stricter: it assigns 9.9 pp fewer \emph{correct} labels and 10.0 pp more \emph{incorrect} labels, while the \emph{incomplete} share stays nearly unchanged.

\begin{table}[ht]
  \centering
  \caption{Human vs. Claude 3.5 label distribution on the 330‑answer test set}
  \label{fig:llm_judge_vs_human}
  \begin{tabular}{lccc}
    & \textbf{Correct} & \textbf{Incomplete} & \textbf{Incorrect} \\
    \hline
    Human evaluation   & 174 (52.7\%) & 63 (19.1\%) & 93 (28.2\%) \\
    LLM‑as‑a‑Judge     & 138 (41.8\%) & 66 (20.0\%) & 126 (38.2\%) \\
    \hline
  \end{tabular}
\end{table}

\subsection{Type I and Type II Risk}
\label{app:typeI-typeII}

We gauge the reliability of the automatic grader using Table \ref{tab:confusion}.

\begin{table}[ht]
\centering
\caption{Confusion matrix (Claude 3.5 vs.\ human labels).}
\label{tab:confusion}
\begin{tabular}{lccc}

 & \multicolumn{3}{c}{\textbf{LLM label}} \\[-2pt]
\cmidrule(lr){2-4}
\textbf{Human label} & Incorrect (0) & Incomplete (1) & Correct (2) \\
\midrule
Incorrect (0)  & 83 & 6  & 4  \\
Incomplete (1) & 25 & 24 & 14 \\
Correct\,(2)   & 18 & 36 & 120 \\
\bottomrule
\end{tabular}
\end{table}

We consider type I error (false accept) when the LLM judges an answer correct (label 2) while the human grader says it is incorrect or merely incomplete, and type II error (false reject) when the LLM judges an answer incorrect (label 0) or incomplete (label 1) while the human grader deems it correct.

With a set of 330 questions:
\[
\begin{aligned}
\text{Type I errors}  &= 4 + 14 = 18 \quad (5.45\%) \\[4pt]
\text{Type II errors} &= 18 + 36 = 54 \quad (16.36\%)
\end{aligned}
\]

The grader is conservative: it underrates correct answers three times more than is over-credits wrong ones. This is desirable in a benchmark where false positives would inflate model scores.

The 16 \% Type II rate means some genuinely correct answers are penalized. Reported accuracies for all models are therefore slightly conservative, a trade-off we accept for fully automated scaling.

\paragraph{Take‑away.}
Despite a small distribution shift, Claude 3.5 aligns with human judgements on more than two‑thirds of \emph{hard} cases, providing a cost‑effective and reproducible grading mechanism for the remainder of our experiments.

\section{Appendix B \\ GHG Emissions Computation Methodology}
\label{app:emissions}

\paragraph{What is at stake?}
Large-scale language models already contribute an electricity demand on par with that of some small nations, and their footprint is growing faster than most mitigation pathways allow.
If research communities omit credible carbon accounting, three risks arise:
\emph{(i) Scientific integrity}: results that overlook energy cost may promote architectures that are infeasible under tightening carbon budgets;
\emph{(ii) Legal exposure}: institutions that publish “AI-for-climate” tools without disclosing their own emissions may soon contravene emerging disclosure laws;
\emph{(iii) Capital-allocation bias}: investors and policymakers, lacking transparent numbers, could funnel resources toward high-footprint solutions that erode limited global carbon budgets instead of toward lower-impact alternatives. Robust footprint measurement is therefore not an optional add-on but a prerequisite for credible, actionable climate-finance research.

\subsection{Why measure the carbon footprint?}
\label{subsec:motivation}

\paragraph{Regulatory and fiduciary context.}
The European \textit{Corporate Sustainability Reporting Directive} (CSRD) and disclosure frameworks such as TCFD\footnote{Task Force on Climate-related Financial Disclosures.} and IFRS~S2\footnote{International Financial Reporting Standards S2.} increasingly require firms and not only heavy industry to quantify Scope 2 (electricity) and Scope 3 (up/down-stream) greenhouse-gas (GHG) emissions. Research groups that propose AI tools for climate finance therefore face a dual responsibility:
(1)~to demonstrate that the use-phase of their models does not undermine the very decarbonisation goals they seek to advance, and (2) to provide transparent numbers that downstream users can incorporate into their own value-chain accounting.

\paragraph{Where do the emissions come from in a RAG pipeline?}
\begin{itemize}
    \item \textbf{Model inference.}  Each forward pass through a large-parameter LLM drives a GPU at dozens to hundreds of watts.
          Although training is more energy-intensive per hour, repeated inference requests dominate in a production search-and-answer workload.  API calls to proprietary back-ends move this energy use off-premises but do not eliminate it.
    \item \textbf{Retrieval and indexing.}  Dense vectorisation of long PDF reports and nearest-neighbour search in FAISS are CPU and memory-heavy and can also leverage GPU if OCR is involved.  While it is a one-off cost in our benchmark, a live system that ingests frequent ESG filings must re-index regularly, making this a non-negligible share of total energy.
    \item \textbf{Evaluation.} Using an LLM-as-a-Judge multiplies the number of model invocations, adding its own emissions line item.
    \item \textbf{Data movement and cooling.}  Every gigabyte shuttled between object storage, RAM and GPU DRAM and every kilowatt-hour dissipated as heat requires additional electricity that is rarely met with 100 \% renewables.
\end{itemize}

\paragraph{Why are the resulting emissions global?}
Computation is executed in hyperscale data-centres scattered across the United States, Europe and Asia,
each connected to a distinct electricity grid with its own carbon intensity. A single Kaggle session may run in Iowa (coal-heavy mix) today and in Belgium (higher share of wind) tomorrow; similarly, an OpenAI or Anthropic request may land on hardware in Oregon, Virginia or Dublin. Consequently, the same Python script yields different real-world emissions depending on where the scheduler places the job and on hourly fluctuations in energy mix.
Embodied carbon in GPUs, network switches and cooling infrastructure further spreads the climate impact across manufacturing hubs in Taiwan, South Korea and the wider semiconductor supply chain.
Quantifying the footprint therefore requires both local energy logs and global carbon-intensity factors, as reflected in the methodology that follows.

\subsection{Study overview}
We distinguish two broad sources of energy consumption in our study:
\begin{enumerate}
  \item \textbf{Local inference runs} (vector‑store creation, RAG pipelines and LLM generation on Kaggle GPUs/CPUs).
  \item \textbf{Remote API calls} to proprietary LLM providers (OpenAI’s GPT‑4o and Anthropic’s Claude 3.5), including the LLM‑as‑a‑Judge evaluation phase.
\end{enumerate}

\subsection{Remote API calls}
For each prompt sent to GPT‑4o or Claude we queried the \textit{EcoLogits}\footnote{\url{https://ecologits.ai/latest/}} service, which
returns a minimum and maximum estimate of GHG emissions (in~kg\,CO\textsubscript{2}\,eq) given the
token counts and the limited public information on vendor infrastructure.  
Because the underlying hardware and regional electricity mixes are opaque, we adopt the maximum value as a conservative upper bound, compute the sample mean~$\bar{e}_{\text{API}}$ and standard deviation~$\sigma_{\text{API}}$ across a calibration batch and then scale by the total number~$N_{\text{API}}$ of queries:

\[
E_{\text{API}}
  = N_{\text{API}} \,\bar{e}_{\text{API}}
  \quad\text{with uncertainty } \pm N_{\text{API}} \,\sigma_{\text{API}}.
\]

\subsection{Local runs (Kaggle)}
\paragraph{Instrumentation.}  
We wrapped every Kaggle notebook in a \texttt{CodeCarbon}\footnote{\url{https://codecarbon.io/}} (v\,\texttt{2.8}) context manager to log:
\begin{itemize}
  \item CPU and RAM energy draw on the host machine;
  \item GPU run‑time in seconds, captured via \texttt{nvidia‑smi}.
\end{itemize}

\paragraph{Converting energy to emissions.}
Let $E_{\text{CPU+RAM}}$ and $t_{\text{GPU}}$ be the cumulative energy (kWh) and GPU run‑time (h) for a given run.  
For each physical host we resolve its ISO country code via \texttt{CodeCarbon}, look up the country‑level carbon‑intensity~$CI$ (kg\,CO\textsubscript{2}\,eq/kWh) from the Climate Change Performance Index  database\footnote{\url{https://ccpi.org}} and compute

\[
E_{\text{local}}
  = CI \,(E_{\text{CPU+RAM}} + t_{\text{GPU}} \times 0.250),
\]
because the NVIDIA P100 on Kaggle is rated at \SI{250}{W} TDP.

\paragraph{Example.}
For a typical 330‑question batch on an unquantized Llama3.1 8B model with a server located in the USA:
\[
\begin{aligned}
E_{\text{CPU+RAM}} &\approx \SI{0.27}{kWh},\\
t_{\text{GPU}} &\approx \SI{4.7}{h},\\
CI_{\text{USA}} &= \SI{0.349}{kg\,CO_2\,eq\,/\,kWh},\\
\Rightarrow\; E_{\text{local}} &\approx (\SI{0.27}{} + 4.7 \times 0.250) \;(\text{kWh}) \\
                       &\quad\;\; \times 0.349 \;(\text{kg\,CO}_2\,eq\,/\,kWh) \\
                       &\approx \SI{0.50}{kg\,CO_2\,eq}.
\end{aligned}
\]

\newpage

\subsection{Emissions summary.}

\begin{table}[ht]
  \centering
  \caption{GHG emissions per \textbf{330-question} run for each model, including evaluation, and for vector-store generation}
  \label{tab:emissions_per_run}
  \begin{threeparttable}
  \begin{tabular}{l
                  S[table-format=2.2]
                  S[table-format=2.2]}
    \toprule
    \textbf{Configuration} &
    {\textbf{Emissions} (kg\,CO\textsubscript{2}\,eq)} &
    {\textbf{Uncertainty} (kg\,CO\textsubscript{2}\,eq)} \\
    \midrule
    LLaMA 3.1-8B (full-precision)         & 0.92 & 0.00 \\
    LLaMA 3.1-8B (4-bit)                  & 0.23 & 0.00 \\
    Mistral NeMo-12B (4-bit)\tnote{\dag}  & 0.38 & 0.00 \\
    GPT-4o (OpenAI API)                   & 2.37 & 1.32 \\
    Claude 3.5 Sonnet (API)               & 2.69 & 1.49 \\
    \midrule
    Vector-store generation               & 0.10 & n/a \\
    \bottomrule
  \end{tabular}
  \begin{tablenotes}[flushleft]
    \footnotesize
    \item[\dag] GPU energy for NeMo was approximated using LLaMA 3.1 power draw; the true value may differ.
  \end{tablenotes}
  \end{threeparttable}
\end{table}

Table~\ref{tab:emissions_per_run} summarises the per-run carbon footprint, including evaluation.

Three patterns stand out:

\begin{enumerate}
  \item \textbf{API calls dominate.}  Remote inference on GPT-4o and Claude accounts for \(\sim\)5 kg CO\(_2\)eq—roughly \SI{77}{\percent} of the subtotal for a single RAG configuration. It is expected because these models are substantialy bigger in size compared to the models we ran locally and each prompt triggers an opaque multi-GPU back-end and must include large uncertainty margins.
  \item \textbf{Quantization pays off.}  Compressing LLaMA 3.1 from full precision to 4-bit slashes emissions from \(0.92\) to \(0.23\;\text{kg\,CO}_2\text{eq}\) (\(\approx75\%\) savings) while retaining answer quality within two percentage points.  Similar gains are expected for NeMo, although its figure is an approximation.
  \item \textbf{Vector stores are not the main emission point.}  End-to-end indexing of all sustainability reports emitted just \(0.10\;\text{kg\,CO}_2\text{eq}\), two orders of magnitude below the API footprint, showing that retrieval costs are negligible compared with repeated LLM inference.
\end{enumerate}

Doubling the experiment to test two RAG pipelines raises the subtotal from \(6.60\) to \(13.19\;\text{kg\,CO}_2\text{eq}\), and the overall benchmark footprint (including index generation) to \(13.29 \pm 5.62\;\text{kg\,CO}_2\text{eq}\).
The error band is driven almost entirely by the vendor-side uncertainty on GPT-4o and Claude. Further transparency from providers about region‑level energy accounting would narrow these bounds and help researchers optimize low-carbon deployments. 

\subsection{Limitations}
\begin{itemize}
    \item Lack of fine‑grained telemetry for GPT‑4o/Claude forces us to rely on coarse upper bounds.
    \item Kaggle does not guarantee the same data‑centre region across sessions, so we approximate with the reported country code for each run.
    \item Carbon‑intensity figures rely on national averages; data‑centre‑specific power‑purchase agreements (PPAs) could yield substantially lower actual emissions.
    \item Provider‑side uncertainty accounts only for hardware variation, not for idle overhead, networking, or cooling system coefficient of performance (COP).
    \item We could not evaluate carbon footprint of Qwen 2.5 and DeepSeek R1 calls because we did not have access to Nebius energy metrics.

\end{itemize}

\clearpage
\section{Appendix C \\ Selected companies and index affiliation}
\label{app:companies}

\begin{center}
\begin{threeparttable}
\begin{tabular}{p{0.28\linewidth} p{0.28\linewidth} p{0.28\linewidth}}
\toprule
\centering\textbf{S\&P500} &
\centering\textbf{CAC40 \& DAX40} &
\centering\textbf{Other}\tabularnewline
\midrule
Apple                       & Axa                     & Alibaba Group \\
AT\&T                       & BNP Paribas             & ArcelorMittal SA \\
Bank of America             & BASF           & Baoshan Iron \& Steel \\
ExxonMobil                  & Engie                   & BP \\
General Electric            & LVMH                    & Hindustan Unilever \\
Alphabet (Google)           & Orange                  & Nestlé \\
Meta Platforms              & Sanofi                  & NTPC \\
Microsoft                   & Siemens AG     & Roche Holding AG \\
NVIDIA                      & Suez                    & Samsung Electronics \\
Pfizer                      & TotalEnergies           & Sinopharm \\
Sysco                       & Veolia Environnement    & SPD Bank \\
\bottomrule
\end{tabular}
\end{threeparttable}
\end{center}

\clearpage
\section*{Appendix D \\ Annotation Guide for Climate Finance Bench}
\label{app:annotation-guide}

\subsection*{1. Introduction}
 
We build a database that links 33 corporate climate reports to

\begin{itemize}
  \item ten analyst-style questions per report,
  \item the reference answers, and
  \item the evidence passages used to derive those answers,
\end{itemize}

so that retrieval-augmented generation (RAG) systems can be evaluated on the same tasks.

Rigorous annotation is critical for trustworthy results.  
This guide sets out the tools, rules, and examples you must follow.  
\textbf{All fields in the database must be written in English.}  
If in doubt, contact your referents before proceeding.

\subsection*{2. Reports to Annotate}

All reports are supplied as PDF files in the shared drive:

\begin{center}
\texttt{Company Reports/}
\end{center}

Verify that every document you use is a climate or sustainability report.

\subsection*{3. Filling in the “Annotation Table” Sheet}

Each annotator completes all columns for their assigned rows, except \texttt{Validity status of the annotation}, which is for the ESG expert only.

\paragraph{Column-by-column instructions}

\begin{enumerate}
  \item \textbf{Annotator’s name}: format: \texttt{firstname\_surname}
  \item \textbf{Company’s name}: exactly the folder name that contains that company’s reports.
  \item \textbf{Fiscal year}
    \begin{itemize}
      \item Use the fiscal year covered, not the publication year.
      \item If the question spans several years, list them: \texttt{2020, 2021}.
      \item Check that the year matches the report contents.
    \end{itemize}
  \item \textbf{Question ID}: one ID for each of the ten questions (e.g.\ \texttt{Q1}).
  \item \textbf{Question}
    \begin{itemize}
      \item Copy exactly from the master question list.
      \item Replace \texttt{FYXXXX} with the concrete year, e.g.\ \texttt{FY2020}.
    \end{itemize}
  \item \textbf{Type of question}: \texttt{PE} (Pure Extraction), \texttt{NR} (Numerical Reasoning) or \texttt{LR} (Logical Reasoning).
  \item \textbf{Answer}
    \begin{itemize}
      \item Write a concise English answer once found.
      \item If the information is not in the report, write exactly:\\
            \texttt{Not available in the retrieved information.}
      \item You may ask internal or external chatbots, but validate the response against the PDF.
    \end{itemize}
  \item \textbf{Documents}
    \begin{itemize}
      \item Use the exact PDF filename(s).
      \item Multiple documents: list in chronological order, separated by commas  
            (e.g.\ \texttt{doc1.pdf, doc2.pdf}).
    \end{itemize}
  \item \textbf{Pages}
    \begin{itemize}
      \item Give the PDF page numbers, not the printed page labels.
      \item One document: \texttt{doc1\{page17, page26\}}.  
            Multiple documents: \texttt{doc1\{page9\}, doc2\{page1, page27\}}.
      \item \texttt{doc1} = first file named in the \textbf{Documents} column, \texttt{doc2} = second file named, etc.
    \end{itemize}
  \item \textbf{Document extracts}
    \begin{itemize}
      \item Copy-paste the full paragraph / table / figure caption.
      \item One extract: \texttt{doc1\{<extract>\}}.  
            Several extracts: \texttt{doc1\{...\}, doc1\{...\}, doc2\{...\}}.
      \item For a table or figure, copy textual content rather than screenshots.
      \item If the PDF is image-only, screenshot, run OCR (e.g.\ ChatGPT), and verify the text.
    \end{itemize}
  \item \textbf{Extract type}: choose from  
        \texttt{text}, \texttt{table}, \texttt{figure}, \texttt{text+table}, \texttt{text+figure}
  \item \textbf{Comments}: pick one of these options
    \begin{itemize}
      \item \texttt{Nothing to report} (default)
      \item \texttt{Uncertain}
      \item \texttt{Additional comments}
    \end{itemize}
  \item \textbf{Additional comments}: fill only if the previous field is \texttt{Additional comments}.  
        Write a short, clear remark.
  \item \textbf{Validity status of the annotation}: ESG expert only.
        Options: \texttt{Validated by the expert} / \texttt{Modified by the expert}.  
        The default entry is \texttt{To be validated}.
\end{enumerate}

\subsection*{4. Annotation Examples}

Examples are provided in the workbook:

\begin{itemize}
  \item \emph{Text answer} → sheet \texttt{Examples / Example 1}
  \item \emph{Table answer} → sheet \texttt{Examples / Example 2}
  \item \emph{Figure answer} → sheet \texttt{Examples / Example 3}
\end{itemize}

\subsection*{5. Quality-Control Procedure}

A dedicated ESG expert reviews all annotations, with special attention to rows whose \textbf{Comments} field is \texttt{Uncertain} or \texttt{Additional comments}.  
The expert then sets the \texttt{Validity status of the annotation} to

\begin{itemize}
  \item \texttt{Validated by the expert}, or
  \item \texttt{Modified by the expert} (if corrections were required).
\end{itemize}

\subsection*{6. Practical Tips}

\begin{itemize}
  \item You may upload the PDF to our internal or external chatbots, ask the question, and then verify the answer in the report.
  \item If the report is an image (no selectable text), take a screenshot, run OCR, and check the transcription carefully.
\end{itemize}

\end{document}